\documentclass[letterpaper,10pt,conference]{ieeeconf}
\IEEEoverridecommandlockouts
\usepackage{graphics}           
\usepackage{times}              
\usepackage{amsmath}            
\usepackage{amssymb}            
\usepackage{graphicx}
\usepackage{algorithm}
\usepackage[noend]{algpseudocode}
\usepackage{booktabs}
\usepackage{color}
\usepackage{listings}
\usepackage{subfiles}
\usepackage{hyperref}
\hypersetup{hidelinks}
\usepackage[nocompress]{cite} % sorts citations automatically, e.g., "[1],[2],[3]" instead of "[2],[3],[1]".
\definecolor{instructioncolor}{rgb}{.5,.5,.5}

\usepackage[font=small]{caption}

\def\figref#1{Fig.~\ref{#1}}

\makeatletter
\usepackage{xspace}
\DeclareRobustCommand\onedot{\futurelet\@let@token\@onedot}
\def\@onedot{\ifx\@let@token.\else.\null\fi\xspace}

\def\etal{{et al}\onedot}
\makeatother

\usepackage{array}
\newcolumntype{L}[1]{>{\raggedright\let\newline\\\arraybackslash\hspace{0pt}}m{#1}}
\newcolumntype{C}[1]{>{\centering\let\newline\\\arraybackslash\hspace{0pt}}m{#1}}
\newcolumntype{R}[1]{>{\raggedleft\let\newline\\\arraybackslash\hspace{0pt}}m{#1}}

\usepackage{bm}
\newcommand{\idle}{\mathrm{idle}}

\newcommand{\ssub}{\mathrm{ss}}
\newcommand{\rmref}{\mathrm{ref}}
\newcommand{\obs}{\mathrm{obs}}

\title{\LARGE\bfseries From Temperature-Limited to Battery-Limited Legged Locomotion}
\author{Rapha\"el S\'egas%
\thanks{This work has been submitted to the IEEE for possible publication. Copyright may be transferred without notice, after which this version may no longer be accessible.}%
\thanks{The authors are with Woven by Toyota, Inc., Tokyo, Japan. Corresponding author: \texttt{raphael.segas@woven.toyota}}
        \and Bal\'azs Opra
        \and L\'eo Ghafari
        \and Thomas Stewart}

\makeatletter
\renewcommand\section{\@startsection{section}{1}{\z@}{1.2ex plus 1ex minus 0.3ex}{0.5ex plus 0.5ex}{\normalfont\normalsize\centering\scshape}}
\renewcommand\subsection{\@startsection{subsection}{2}{\z@}{1.2ex plus 1ex minus 0.3ex}{0.5ex plus 0.5ex}{\normalfont\normalsize\itshape}}
\makeatother

\begin{document}
\thispagestyle{empty}
\pagestyle{empty}
\IEEEaftertitletext{\vspace{-4mm}}
\maketitle
% \vspace{-3mm}

\begin{abstract}
On legged robots, sustained locomotion often ends in thermal shutdown rather than battery depletion: the heat of walking concentrates in a few joint motors, which overheat long before the battery is empty. We train thermally aware locomotion policies by reinforcement learning without ever observing temperature, and demonstrate this on a Unitree Go2 quadruped. We first identify a coupled 13-node lumped RC thermal network, fitted on two runs of a single robot and calibrated against the measured
electrical power, that predicts all twelve motor temperatures with a mean RMSE below 1\,\textdegree C across different robot instances, gaits, payloads, and speeds. Under a sustained gait, the temperatures the motors will settle at follow directly from the gait-averaged squared torques and joint speeds. Our core contribution is a reward that drives down the hottest predicted asymptotes: thermal time constants exceed the credit-assignment horizon of reinforcement learning by orders of magnitude, but predicted asymptotes respond to torque and joint speed within a single gait cycle, so the policy learns to lower and spread heat with no temperature input. On the robot, the policy turns sustained locomotion from temperature-limited to battery-limited: with a 5\,kg payload at 1\,m/s, a learned baseline reaches thermal shutdown after 19 minutes (1.1\,km), whereas our policy runs the battery down after 174 minutes (10.0\,km) with its hottest motor 14\,\textdegree C below the shutdown threshold in a 26\,\textdegree C room; on an unloaded factory circuit with stairs and ramps, the baseline overheats after 25 laps (6.2\,km), while our policy completes 47 (11.2\,km) on one battery with 22\,\textdegree C of margin. 
\end{abstract}

% Keep this early top float beside the abstract in the first-page right column.
\begin{figure}[!t]
\centering
\includegraphics[width=0.92\columnwidth]{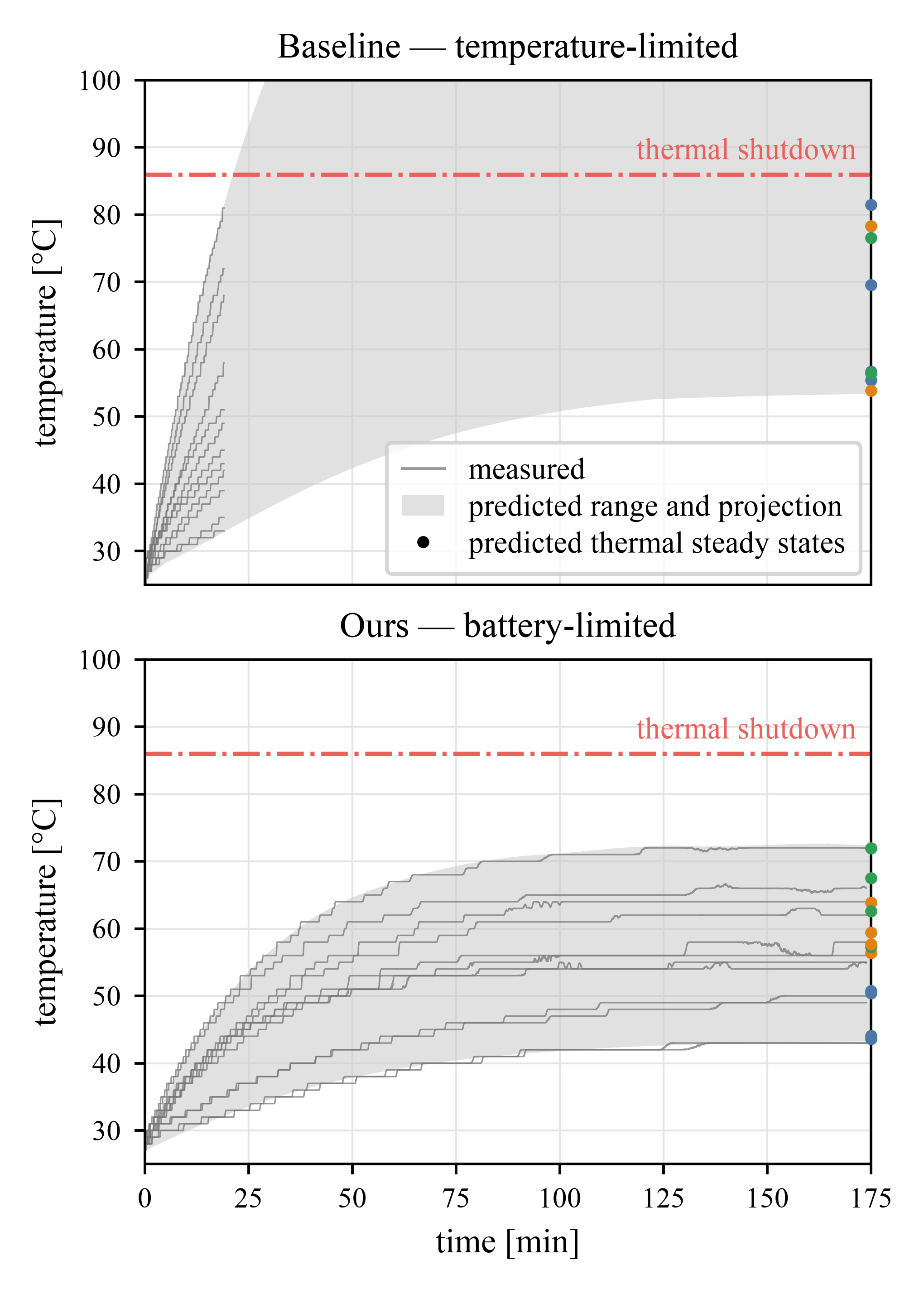}\par
\vspace{2pt}
% Photos are 2100 x 1575 at 72 dpi; trim is left, bottom, right, top in bp.
% Matching 4:3 crops retain the full robot and show the payload mounting closely.
\includegraphics[width=0.495\columnwidth,trim={300bp 235bp 220bp 155bp},clip]{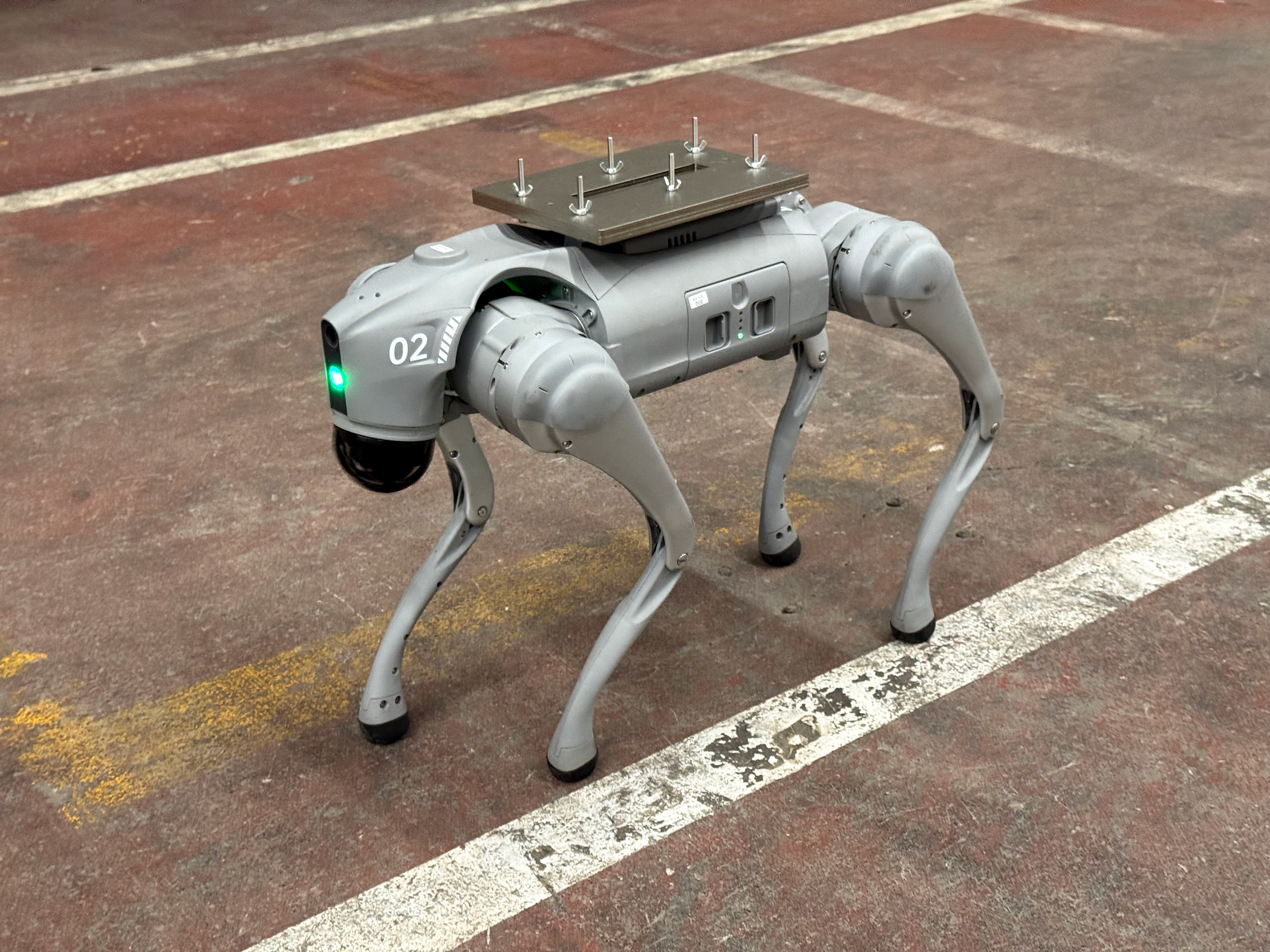}\hfill
\includegraphics[width=0.495\columnwidth,trim={255bp 210bp 345bp 240bp},clip]{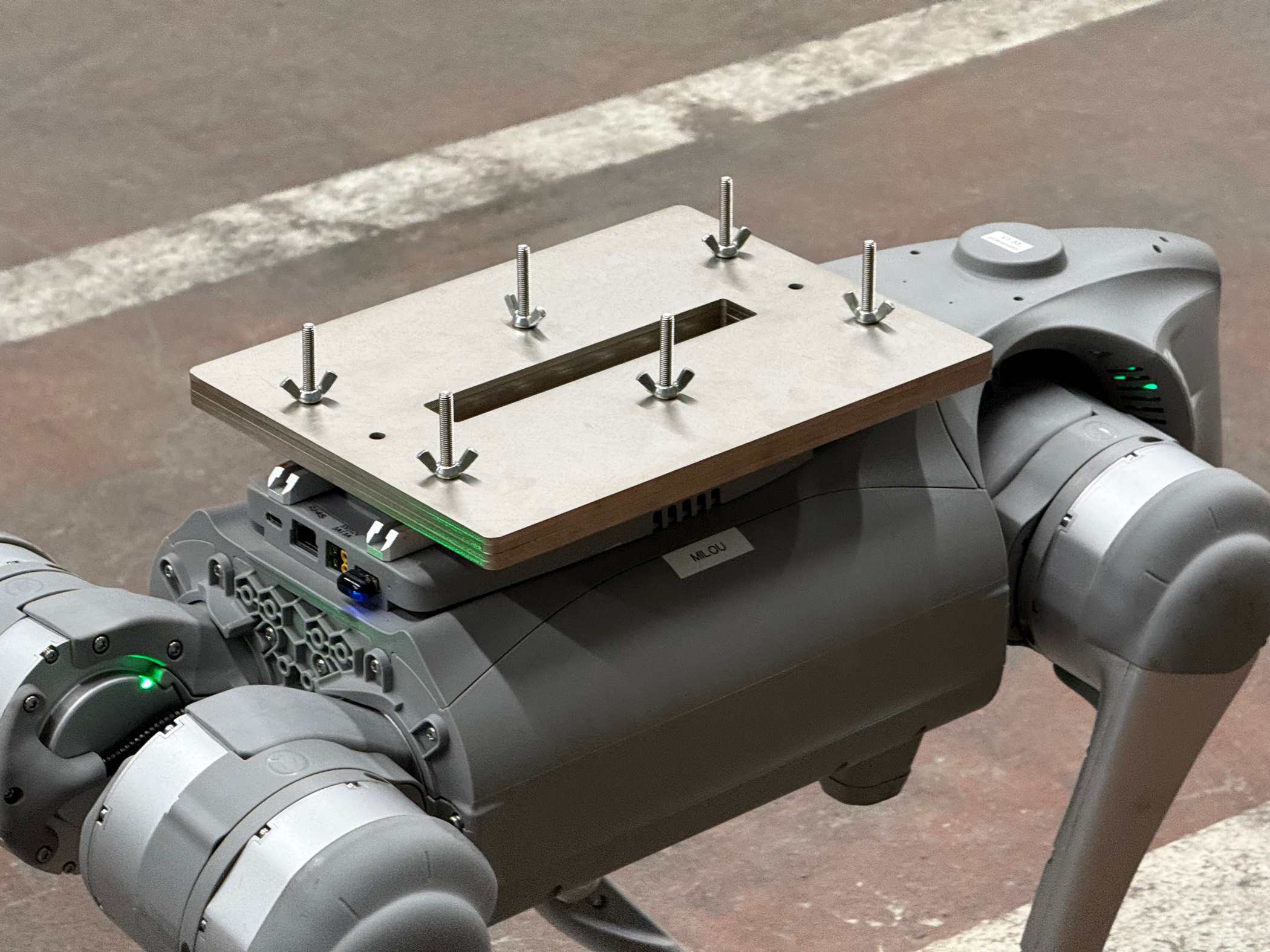}
\par\vspace{0mm}
\caption{Sustained 1\,m/s walking with a 5\,kg payload. The baseline reaches thermal shutdown after 19 min (top); our thermal policy drains the battery after 174 min with 14\,\textdegree C of headroom (middle). Grey: the 12 measured motor temperatures; shaded: min–max of our thermal model, projected with constant gait inputs; dots: predicted asymptotes (blue hip, orange thigh, green calf), those above the axis omitted. Bottom: the payload and its mounting.}
\label{fig:baseline-vs-ours}
\vspace{-1mm}
\end{figure}

\section{Introduction} \label{sec:intro}

%% WHY
Learned controllers have made legged robots agile and robust~\cite{hwangbo2019scirob,lee2020scirob,miki2022scirob,rudin2022corl,nahrendra2023icra}: they climb stairs and cross unstructured terrain on proprioception alone~\cite{lee2020scirob,nahrendra2023icra}. With this surge in capability, these robots are increasingly asked to perform long-horizon tasks: patrolling a factory, running an inspection circuit, or carrying parcels up the stairs of an apartment block, often under heavy payloads. In practice, such runs often end long before the battery is empty, because a joint motor overheats. Electric legged robots are typically driven by permanent magnet synchronous motors~(PMSMs) whose dominant loss, Joule heating, grows with the square of the torque~\cite{wallscheid2016tec,bjelonic2026ijrr}. In a compactly packaged leg, that heat escapes only slowly. Nahrendra \etal~\cite{nahrendra2023icra} had to slow their robot down in summer trials to keep its front-leg motors from overheating, Matsuzawa \etal~\cite{matsuzawa2026arxiv} found heat concentrating in the hip joints of their legged-wheeled robot until the load was redistributed, and the humanoid that beat the human half-marathon record in April 2026 ran on liquid-cooled leg motors~\cite{de2026spectrum}, a remedy that adds mass, power draw, and cost. A stock Unitree Go2 walking at 1\,m/s in a 27\,\textdegree C room shuts down after about two hours (7.7\,km), with charge to spare and no payload. In a 22\,\textdegree C room, a learned policy trained without thermal awareness, the baseline of Sec.~\ref{sec:baseline-ours}, lasts 19 minutes (1.1\,km) with 5\,kg on its back (\figref{fig:baseline-vs-ours}).

%% PROBLEM
In this paper, we address the problem of making learned locomotion policies thermally aware, so that sustained locomotion is limited by the battery rather than by motor temperatures. Two properties set this apart from energy minimization. First, overheating is a distribution problem: the robot shuts down when \emph{any} motor crosses its threshold (around 86\,\textdegree C on the Go2), so what matters is the hottest joint, not the sum over joints that an energy objective minimizes~\cite{bjelonic2026ijrr}. A gait can be efficient and still thermally fatal when it loads a single joint, a pattern Lin \etal~\cite{lin2025robot} observed on a quadruped where a few joints overheated while the others stayed cool. Second, thermal time constants are tens of minutes, whereas the Joule heating that primarily drives them follows torque within a control step. Over a reinforcement learning~(RL) episode of a few seconds, temperatures barely change, making credit assignment of temperature-based rewards exceedingly challenging.

%% HOW & WHAT
Thermal limits are typically handled in three ways. Hardware cooling pushes the thermal envelope~\cite{sevinchan2018ate,zhu2021icar} at the price of mass and power. Model-based control enforces temperature limits through torque caps, thermally constrained trajectory optimization, or recovery postures~\cite{guilbert2008ijrr,trujillo2009icra,kumagai2014humanoids,tan2018taros,jorgensen2019ral,kawaharazuka2020ral}, mostly on fixed-base or slowly moving systems where capping torque does not endanger balance. In learned locomotion, the issue has often been addressed by heuristics, such as the power distribution term of Nahrendra \etal~\cite{nahrendra2023icra} that penalizes the variance of mechanical power across joints and, more recently, by feeding motor temperatures to the policy as observations with a reward derived from a control barrier function~(CBF)~\cite{ames2019ecc} on the temperature derivative~\cite{muller2026ral,qian2026arxiv}. M\"uller \etal~\cite{muller2026ral} keep the neck actuator of an animated character below its limit with a decoupled first-order thermal model, and Qian \etal~\cite{qian2026arxiv} combine the same reward with the whole-body thermal network of Lin \etal~\cite{lin2025robot} to carry 3\,kg on a Unitree A1 for 27 minutes, up from 7 without the reward. Both grapple with the long thermal time scale. Qian \etal~\cite{qian2026arxiv} note that the CBF term ``typically exhibits limited variation'' within an episode, initialize temperatures in a band around the threshold to obtain a learning signal, and report that the policy ``tends to adopt conservative walking postures even when motor temperatures are low'', which hinders performance on slopes and stairs. Wan \etal~\cite{wan2026arxiv} soften this with a residual policy conditioned on the thermal state, at the cost of a second training stage; the credit-assignment difficulty reported by Qian \etal remains, since the thermal state barely moves within an episode and the learning signal can only come from its variation across episodes.

We take a different route and do not observe temperature at all. Our key observation is that, under sustained locomotion, the thermal steady state the motors settle into is set by a single gait cycle: once a coupled thermal model of the robot is identified, the temperatures at which heating and cooling equilibrate follow from the gait-averaged squared torques and joint speeds via one small linear solve. Unlike the temperatures themselves, these predicted asymptotes respond to a change in torque or joint speed within a single gait cycle. This collapses the credit-assignment horizon from thermal to gait time scales, leaves the policy input unchanged, and keeps evaluation independent of the initial thermal state. We first identify a compact, coupled 13-node lumped RC thermal network of the Go2, anchored in the measured electrical power balance so that fitted parameters are physical, and we turn it into a reward by penalizing a soft maximum of the predicted asymptotes, which lowers the hottest joints and spreads the load across the others instead of pulling down a sum. We incorporate this reward into a strong blind locomotion pipeline~\cite{rudin2022corl,nahrendra2023icra,bjelonic2026ijrr} and train policies using proximal policy optimization~(PPO)~\cite{schulman2017arxiv} on diverse terrains including stairs and slopes, so the thermal margin does not come at the expense of terrain traversability. Nothing in our approach is specific to blind policies, quadrupeds, or the on-policy algorithm we use.

%% MAIN CONTRIBUTION
The main contribution of this paper is a steady-state-aware thermal reward that makes learned locomotion policies thermally aware without temperature observations: it penalizes a soft maximum of the thermal asymptotes predicted from gait-averaged squared torques and joint speeds through a single linear solve of an identified thermal network, and thereby incentivizes the policy to lower and spread the thermal load across joints. Two further contributions support it: a measurement-first identification methodology that pins the quiescent power, the Joule heating coefficient, and the temperature coefficient of resistance electrically before any temperature is fitted, so that the parameters stay physical; and a compact 22-parameter, 13-node thermal network of the Unitree Go2, identified from two runs of a single robot and reported in full (Tab.~\ref{tab:parameters}).

%% KEY CLAIMS
We make five claims.
(i)~Our identified network predicts all twelve motor temperatures of a Go2 with a mean RMSE below 1\,\textdegree C on runs it was not fitted on, including a second robot, learned gaits, payloads up to 10\,kg, speeds up to 1.4\,m/s, sustained walks, and stop-and-go runs.
(ii)~Electrical anchoring keeps the identified parameters physical, whereas a temperature-only fit of the same model drifts to nonphysical values and generalizes worse.
(iii)~Our reward lowers the hottest predicted thermal asymptote at every payload in simulation, and its weight sets the margin.
(iv)~The resulting policy retains terrain traversal capability over steep stairs and slopes, in simulation and on a factory circuit.
(v)~Without temperature observations, it shifts Go2 locomotion from temperature-limited to battery-limited operation, both with a 5\,kg payload on flat ground and unloaded on the same circuit.
Sec.~\ref{sec:experiments} supports the claims in this order: (i) and (ii) on twenty-one held-out runs, (iii) and (iv) in simulation, and (iv) and (v) on the robot.

\section{Methodology}

% Our methodology is simple and reproducible: we first identify a compact thermal model that predicts the twelve motor temperatures, then add a reward that brings down the thermal steady states it predicts.

\subsection{Thermal system identification}

\begin{figure*}[t]
\centering
\includegraphics[width=0.90\textwidth]{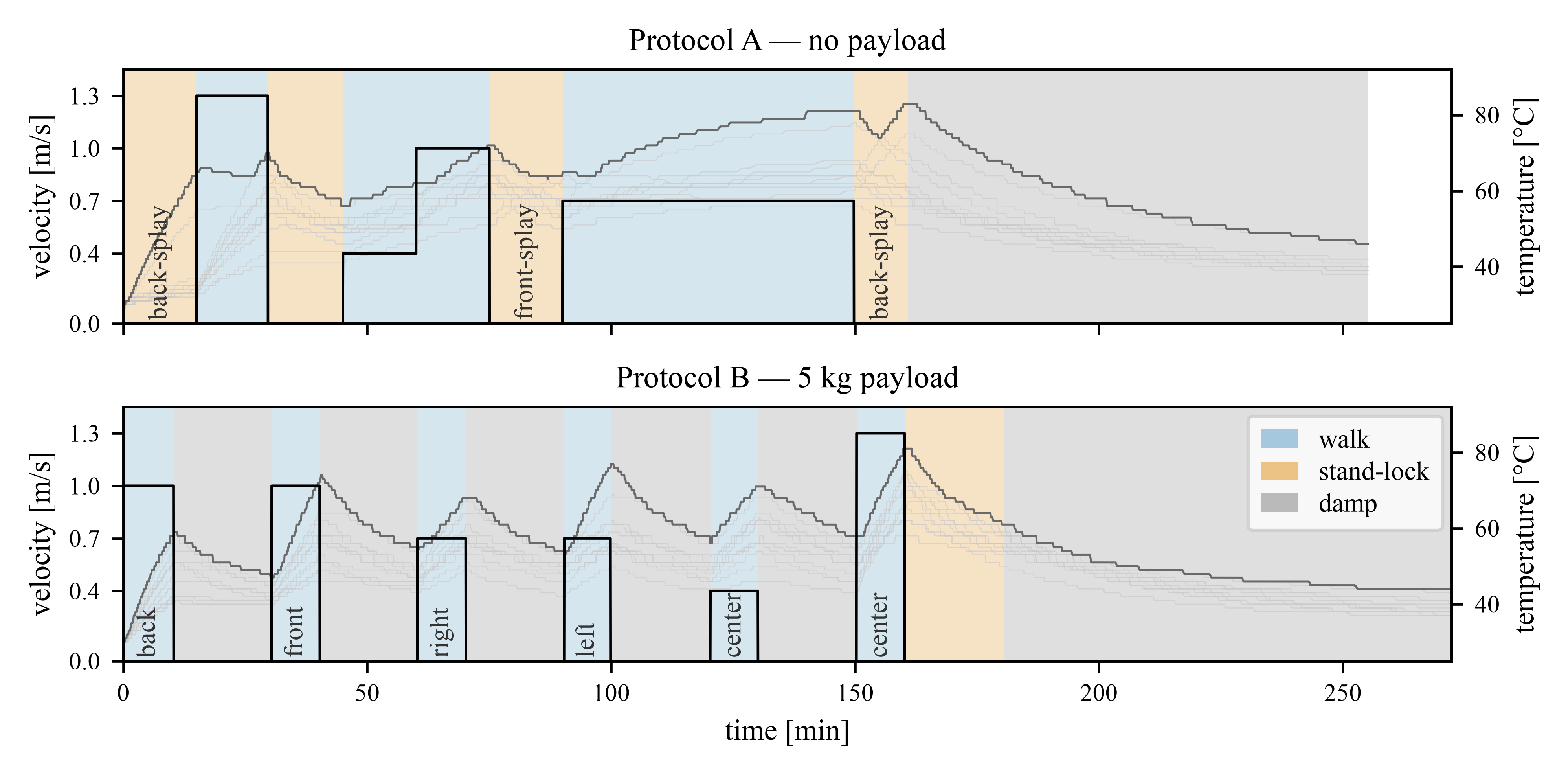}
\par\vspace{-3mm}
\caption{The two training runs, following protocols A and B. The twelve temperatures (light-grey solid lines) are recorded every two seconds. Segments are designed and arranged to produce a large range of temperatures while keeping their maximum (dark-grey solid line) under Go2's thermal shutdown threshold of about 86\,\textdegree C.}
\label{fig:protocols}
\vspace{-1mm}
\end{figure*}

\subsubsection{Experimental protocols} \label{sec:exp-protocol}
We conduct all experiments indoors to prevent wind and direct sun exposure, in a large flat 60 by 25\,m room, so that straight stretches dominate the data. We ensure that the robots are at room temperature before every recording. We sample all motor temperatures, electrical quantities, joint torques and velocities at 50\,Hz and log their statistics every 2\,s. We record ambient temperature at the same cadence with a Sensirion SHT40 sensor ($\pm0.2$\,\textdegree C), and measure robot velocity independently with a stopwatch.

We design two experimental protocols (Fig.~\ref{fig:protocols}). Each is a series of distinct segments spanning forward commanded velocities, payloads, and static stances:
\begin{itemize}
\item protocol A consists of a sweep of forward commanded velocities (over 0.4--1.3\,m/s) and static stances via the built-in stand-lock mode, without any payload on the robot. Before every static segment, we briefly lift the robot off the ground to reset its stance. In the back- and front-splayed static stances, the rear and front legs are pulled apart by an extra $5\pm0.5$\,cm, respectively.
\item protocol B has the robot carry a 5\,kg payload on its back. We shift the payload's center of mass~(CoM) by $\pm8$\,cm fore-aft and $\pm3.2$\,cm laterally and sweep the commanded velocity over 0.4--1.3\,m/s, with a 20-minute damp rest in between each walk.
\end{itemize}
Each protocol concludes with a 90-minute damp segment where the robot rests on a plastic stand under its torso, so that only its feet touch the ground. We do not rest the robot directly on the ground, to avoid an additional torso-to-ground conduction path.

Our training set for fitting the thermal model consists of two runs of over four hours (one per protocol) of a single robot and the stock Unitree locomotion controller (Unitree's sport mode, commanded through the Motion Services Interface V2.0 of the Go2 SDK) for ease of reproducibility by the community.

Segments, individually or in combination, help identify key thermal and electrical parameters introduced in Sec.~\ref{sec:thermal-model} and Appendix~\ref{app:node-level}: damp phases give the electrical quiescent power directly, the splayed static stances predominantly excite hip nodes and their cool/hot pair helps identify the temperature coefficient of the resistance, shifted payloads break symmetries, the two payload levels (0 and 5\,kg) disambiguate secondary heating terms, and the 60-minute sustained walk anchors the thermal asymptotes while the velocity sweep sets the forced-convection terms.

We additionally record nine validation runs, used for cross-validation and structure selection (Sec.~\ref{sec:crossval}), and twelve held-out test runs, scored once with the final model. Together they cover about 39 hours on two different Go2 robots (we fit on one of them), three controller families (the stock controller, DreamWaQ baselines of Sec.~\ref{sec:baseline-ours}, and checkpoints of our thermal policy), payloads from 0 to 10\,kg including off-center loads, body speeds from 0.4 to 1.4\,m/s, sustained walks on flat ground (Sec.~\ref{sec:flat-endurance}) and on the factory circuit (Sec.~\ref{sec:factory-circuit}), and stop-and-go runs with stand-lock or damp phases in between.

\subsubsection{Thermal model} \label{sec:thermal-model}

We propose a compact 22-parameter model: a coupled 13-node lumped RC thermal network anchored by an electrical power balance.

\emph{Notations.} Joints are indexed $j=(\ell,y)$ with leg $\ell\in\{\mathrm{FR,FL,RR,RL}\}$ and type $y\in\{\mathrm{hip,thigh,calf}\}$. We denote $u_j=\tau_j^2$ and $a_j=|\dot q_j|$ where $\tau_j$ is the joint-side current-based torque estimate and $\dot q_j$ the joint velocity. Since the calf motors have an additional gear ratio of $G\approx1.917$, we define the constants $r_j=G^{-2}$ and $\gamma_j=G$ for calf joints ($r_j=\gamma_j=1$ otherwise) to convert joint-side quantities to the motor side. $v$ is the measured body speed and $T_a$ the ambient temperature. The state $T=(T_1,\dots,T_{12},T_B)\in\mathbb{R}^{13}$ consists of twelve sensed stator/case lumps and one hidden body node.

\emph{Heat sources.} Each actuator node receives a Joule heating term $P^{J}_j$ (net of a bypass share to the body), a speed-proportional friction loss with coefficient $f$, and its quiescent dissipation $p_{\idle}$; the body node receives the bypass heat and the board quiescent power $p^{B}_{\idle}$:
\begin{equation}
P^{J}_j = k_m\, r_j\, u_j\,\bigl(1+\alpha\,(T_j-T_{\rmref})\bigr),
\label{eq:copper}
\end{equation}
\begin{equation}
\begin{gathered}
Q_j = \bigl(1-\varphi_{y(j)}\bigr)\,P^{J}_j + f\,\gamma_j\,a_j + p_{\idle},
\\
Q^{B} = \sum_{j=1}^{12} \varphi_{y(j)}\,P^{J}_j \;+\; p^{B}_{\idle},
\end{gathered}
\label{eq:sources}
\end{equation}
where $k_m$ and $\alpha$ are the Joule heating coefficient and temperature coefficient of resistance, respectively; the bypass shares $\varphi_y$ (one per joint type) route part of each actuator's Joule heat directly into the body rather than the sensed lump; $T_{\rmref}$ is equal to 25\,\textdegree C by convention (not a free parameter).

\emph{Network dynamics.} Appendix~\ref{app:node-level} details the node-level equations. We stack them in a matrix-vector form:
\begin{equation}
\begin{gathered}
\mathbf{C}\,\dot T \;=\; Q(u,a,T)\\
\;-\;\mathbf{G}_a(v)\,\bigl(T-T_a\mathbf{1}\bigr)
\;-\;\mathbf{L}\,T,
\\
\hat T_j = \mathrm{lag}\bigl(T_j;\tau_{\obs}\bigr),
\end{gathered}
\label{eq:network}
\end{equation}
where $\mathbf{C}=\mathrm{diag}(C_m\mathbf{1}_{12},\, C_B)$ is the diagonal matrix of thermal capacities; $\mathbf{G}_a(v)$ is the diagonal matrix of ambient conductances (modeled with a power-law forced-convection term for actuator nodes); $T_a$ is the ambient temperature; $\mathbf{L}$ is the weighted graph Laplacian of the conduction topology; and the measured temperatures $\hat T_j$ follow the node states through a first-order lag.

\subsubsection{Electrical power balance} \label{sec:elec-balance}
Fitting temperatures alone cannot fix the scale of the parameters: multiplying the thermal capacities, all conductances, and all heat sources in \eqref{eq:network} by the same factor leaves the temperature trajectories unchanged.
Hence, we consider an electrical power balance to anchor fitted parameters to physical values:

\begin{equation}
\begin{aligned}
P_{\mathrm{bus}} &\;=\; p^{\mathrm{tot}}_{\idle}\\
&\;+\; \sum_{j=1}^{12} k_m\, r_j\, u_j\,\bigl(1+\alpha\,( \hat T_j -
T_{\rmref})\bigr)\\
&\;+\; \sum_{j=1}^{12} \tau_j\,\dot q_j,
\end{aligned}
\label{eq:balance}
\end{equation}
where $P_{\mathrm{bus}}=V\,|I|$ is the bus power and $p^{\mathrm{tot}}_{\idle}=12\,p_{\idle} + p^{B}_{\idle}$ the total quiescent power. Note that the mechanical term is \emph{net and signed} (braking energy returns to the bus). Since torque $\tau_j$ is a current-based estimate, friction is contained in the mechanical term and does not appear separately.

\subsubsection{Fitting and structure selection} \label{sec:crossval}

We fit in two stages (Appendix~\ref{app:sys-id}): the segment-averaged balance \eqref{eq:balance} gives $p^{\mathrm{tot}}_{\idle}$ directly from the damp segments, then $k_m$ and $\alpha$ by weighted linear least squares on the stand and walk segments; with these three constants fixed, the remaining 19 parameters follow from nonlinear least squares on the displayed temperatures of both training runs. We select the network structure by leave-one-run-out~(LORO) cross-validation on the combined training and validation runs, tracing a compactness frontier (Fig.~\ref{fig:frontier}) of validation error against parameter count, and refit the selected structure on the two training runs only.

\subsection{Steady-state-aware thermal rewards} \label{sec:tss-reward}

Thermal time constants are tens of minutes, far longer than an RL episode and the horizon over which PPO assigns credit: a 10\,W change in a motor's Joule heating moves its temperature by about half a degree over 20\,s, while its asymptote moves by roughly twenty degrees. A reward on the temperature level thus barely distinguishes actions that increase heating from those that reduce it, and a reward on its derivative~\cite{qian2026arxiv} conflates the quality of the gait with the thermal state the episode started in, since the same gait heats fast from cold and not at all near equilibrium. The asymptote, in contrast, is an instantaneous function of the gait alone, and we penalize a soft maximum of the asymptotes predicted by the identified model.

Since all terms on the right-hand side of \eqref{eq:network} are affine in $T$, we can rewrite it in the compact form:
\begin{equation}
\begin{gathered}
\mathbf{C}\,\dot T \;=\; -\mathbf{A}(u,v)\,T\;+\;b(u,a,v,T_a),
\\
\hat T_j = \mathrm{lag}\bigl(T_j;\tau_{\obs}\bigr).
\end{gathered}
\label{eq:compact}
\end{equation}

\emph{Steady state.} Under sustained locomotion, we decompose squared torque and absolute joint velocities into gait-averaged and residual terms:
\begin{equation}
u\;=\;\bar u\;+\;\delta u,
\quad
a\;=\;\bar a\;+\;\delta a.
\end{equation}
The network \eqref{eq:compact} acts as a low-pass filter: residual terms (with frequencies greater than 1 Hz) are attenuated by a factor greater than $2\cdot10^{3}$. In the long run, setting $\dot T=0$, we then obtain a linear system for the thermal steady states $T_{\ssub}$:
\begin{equation}
\begin{gathered}
\mathbf{A}(\bar u, v)\;T_{\ssub} \;=\; b(\bar u,\bar a,v,T_a),
\\
T_{\ssub} \;=\; \mathbf{A}^{-1}b.
\end{gathered}
\label{eq:ss}
\end{equation}
Solving this $13\times13$ system costs microseconds, cheap enough to evaluate at every step of every environment. The capacities and the observation lag only affect thermal transients and drop out of \eqref{eq:ss}, so the map from gait to asymptotes depends on only 19 of the 22 parameters.

During RL training, we approximate gait-averaged terms by an exponential moving average~(EMA) over a 1-second time horizon. We replace the measured body speed $v$ by the commanded velocity $v^{\mathrm{cmd}}$ to prevent reward hacking, and set $T_{a}=30^\circ\mathrm{C}$.
We only keep the first 12 entries of the solution (body node excluded), and normalize the vector:
\begin{equation}
T^{*}_{\mathrm{ss}}\;=\;\max\!\left(0,\;
\frac{T_{\ssub,\,1:12} - T_{\mathrm{lo}}}{T_{\mathrm{hi}} - T_{\mathrm{lo}}}
\right),
\label{eq:Tss*}
\end{equation}
with $T_{\mathrm{lo}}\;=\;40^\circ\mathrm{C}\;$ and $\;T_{\mathrm{hi}}\;=\;80^\circ\mathrm{C}$.

The steady-state-aware thermal reward is defined as:
\begin{equation}
r_{\mathrm{th}}\;=\;\left\lVert T^*_{\mathrm{ss}} \right\rVert_p,
\label{eq:Tss-reward}
\end{equation}
with $\left\lVert x \right\rVert_p=(\tfrac1{12}\sum_i x_i^p)^{1/p}$ the $p$-th power mean on $\mathbb{R}^{12}$.

$p$ interpolates between a simple average ($p=1$) and a maximum over $T^*_{\mathrm{ss}}$ ($p=+\infty$). Large but finite $p$ values ($\geq 20$) concentrate the penalty on the hottest joints while spreading the gradient signal over them (a smoothing property $p=+\infty$ lacks). We use $p=20$ in the remainder of this work. In contrast, a variance penalty would only equalize the asymptotes and can be lowered by heating the cool joints.

\subsection{Baselines and our thermal policy} \label{sec:baseline-ours}

\emph{Baseline.} We perform dynamics system identification following the PACE methodology~\cite{bjelonic2026ijrr} to bridge the sim-to-real gap (PD gains are not randomized). In simulation, we train blind locomotion policies in Isaac Lab on diverse procedurally generated terrains, including open-tread and solid stairs with risers up to 15 and 20\,cm, respectively. We follow the game-inspired, performance-adaptive terrain curriculum of Rudin~\etal~\cite{rudin2022corl} with 10 difficulty levels per terrain variant. We randomize the payload mass added to the torso over 0--5\,kg, as well as the torso's fore-aft CoM offset over $\pm5$\,cm to improve sim-to-real robustness. The baseline is a re-implementation of DreamWaQ~\cite{nahrendra2023icra}. We train each policy with PPO for up to 15,000 learning iterations, each comprising 24 rollout steps per environment followed by a policy and estimator update.

\emph{Our thermal policy.} Our thermal policy keeps the DreamWaQ reward terms and weights, adds the steady-state-aware thermal reward of Sec.~\ref{sec:tss-reward} with weight $w_{\mathrm{th}}=-0.2$, and drops three terms that it renders redundant. Power distribution and joint power are both proxies for the heat that the thermal reward penalizes directly; body height is unnecessary because supporting weight on straighter legs lowers calf torque, so a high-body stance emerges on its own. We switch the thermal reward on, at full weight, after 5,000 iterations once terrains are already solved.

\emph{Strengthened baselines.} As a stronger point of comparison we also train the baseline with an added squared-torque penalty of weight $-3\cdot10^{-4}$, the largest weight at which policies still learn to walk on all terrains (strengthened baseline). Strengthening the power-distribution or joint-power terms instead performed worse in the gait evaluation (Sec.~\ref{sec:gait-eval}) and is not reported.

\begin{table}[t]
\vspace{1.5mm}
\caption{Performance of the thermal model and its ablations: RMSE in \textdegree C of the displayed motor temperatures over all 12 joints (all) and of the maximum over the twelve joints (hot), averaged over runs. Train: the two training runs (fit); val: the nine validation runs; test: twelve held-out runs scored once. $k$ counts all identified parameters (electrical + thermal). Ablations 2--5 are the best structure of their family selected by LORO.}
\label{tab:thermal-performance}
\centering
\footnotesize
\setlength{\tabcolsep}{3.5pt}
\begin{tabular}{@{}l c rr rr rr@{}}
\toprule
Model & $k$ & \multicolumn{2}{c}{Train} & \multicolumn{2}{c}{Val} & \multicolumn{2}{c}{Test} \\
\cmidrule(lr){3-4}\cmidrule(lr){5-6}\cmidrule(lr){7-8}
 & & all & hot & all & hot & all & hot \\
\midrule
22-parameter model (ours) & 22 & 0.78 & 0.74 & 0.88 & 0.85 & 0.97 & 0.85 \\
\midrule
1: no electrical identification & 22 & 0.77 & 0.71 & 0.90 & 0.88 & 1.04 & 0.98 \\
2: no joint-velocity heating & 21 & 0.79 & 0.76 & 1.00 & 1.07 & 1.25 & 1.56 \\
3: no bypass share to body & 21 & 0.95 & 1.15 & 1.35 & 1.66 & 1.83 & 2.57 \\
4: no quiescent power terms & 20 & 1.14 & 1.09 & 1.68 & 2.02 & 1.87 & 1.82 \\
5: no calf--thigh conductance & 21 & 1.60 & 1.60 & 2.69 & 4.10 & 3.33 & 6.68 \\
\bottomrule
\end{tabular}
\par\vspace{-2mm}
\end{table}

\begin{figure}[t]
\centering
\includegraphics[width=0.94\linewidth]{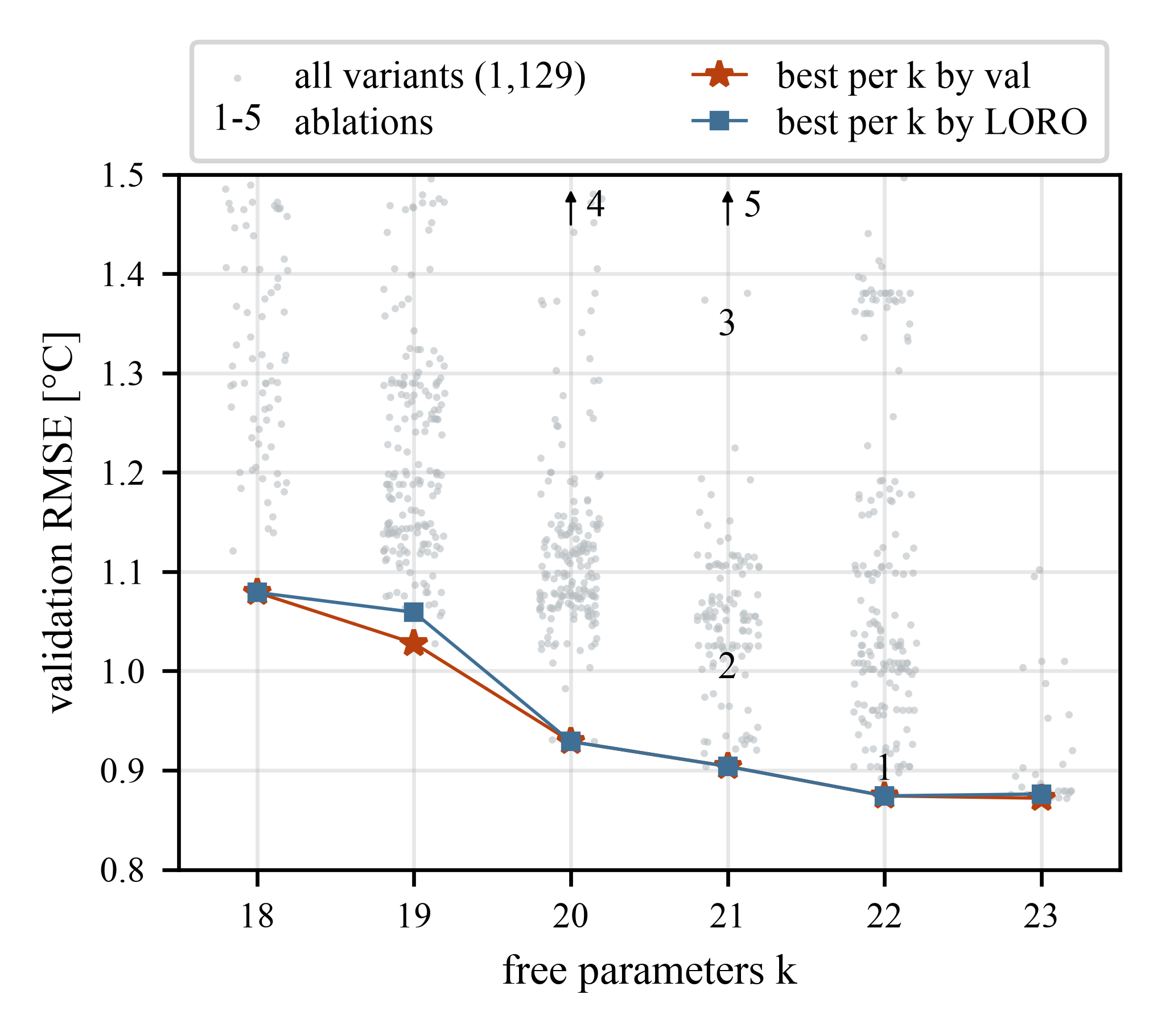}
\par\vspace{-2.5mm}
\caption{Compactness frontier over 1,129 structural variants of the network, scored on the validation runs. Red: best structure per parameter count by validation RMSE; blue: by LORO. Numbers mark ablations 1--5 from Tab.~\ref{tab:thermal-performance}; arrows point off-scale.}
\label{fig:frontier}
\vspace{-1mm}
\end{figure}

\section{Experiments} \label{sec:experiments}

The main focus of this work is a reward that makes learned locomotion policies thermally aware without observing temperature. Our experiments support the five claims of Sec.~\ref{sec:intro} in order. Sec.~\ref{sec:sysid-exp} scores the identified network on twenty-one held-out runs (i) and compares it with a temperature-only fit (ii). Sec.~\ref{sec:gait-eval} isolates the reward in simulation, where payload and ambient are controlled (iii). Sec.~\ref{sec:terrain} evaluates terrain traversal in simulation (iv). Sec.~\ref{sec:flat-endurance} and~\ref{sec:factory-circuit} take the deployed policy to the robot, with a 5\,kg payload on flat ground and unloaded on the factory circuit (iv, v).

\subsection{Thermal system identification} \label{sec:sysid-exp}

Tab.~\ref{tab:thermal-performance} reports the RMSE of the displayed motor temperatures on the training, validation, and test runs, over all twelve joints and for their maximum (hot), the quantity that triggers shutdown. The twenty-one held-out runs average below 1\,\textdegree C, and no single run exceeds 1.25\,\textdegree C over the twelve joints or 1.65\,\textdegree C on the maximum temperature, including the ten runs that reach thermal shutdown and the two that end in battery depletion. The same table lists five ablations. Fitting the same 22 parameters from temperature alone (ablation 1) generalizes worse and, more importantly, drifts to nonphysical values (quiescent power fitted at 18\,W against 44\,W measured, $\alpha$ at $0.47\times$ copper, $k_m +24$ \%, friction $-50$ \%, body-to-ambient conductance $-72$ \%, body capacity $-41$ \%), which is evidence for claim (ii). Removing either the joint-velocity heating term or the bypass share to the body (ablations 2 and 3) has a sizable effect on accuracy, a hint that a torque-only penalty does not suffice for thermal awareness. Removing quiescent powers or the calf-thigh conductance (ablations 4 and 5) deteriorates accuracy sharply. Fig.~\ref{fig:frontier} places the five ablations on the compactness frontier of the 1,129 structures we screened: validation error improves up to 22 parameters and then plateaus, and every ablation lies above the frontier. Appendix~\ref{app:sys-id} lists all 22 identified parameters; the calf--thigh edge has the largest conduction edge, consistent with the calf motor sitting beside the thigh motor on the Go2, as well as the poor score of ablation 5.

\subsection{Gait evaluation in simulation} \label{sec:gait-eval}

Under a 1\,m/s forward commanded velocity, we roll out 40-second trajectories on flat terrain for payloads ranging from 0 through 5\,kg in 1\,kg increments. We discard the first five seconds, record the gait statistics, and predict the thermal asymptotes with our model.

\begin{figure}[t]
\vspace{1.5mm}
\centering
\includegraphics[width=0.95\columnwidth]{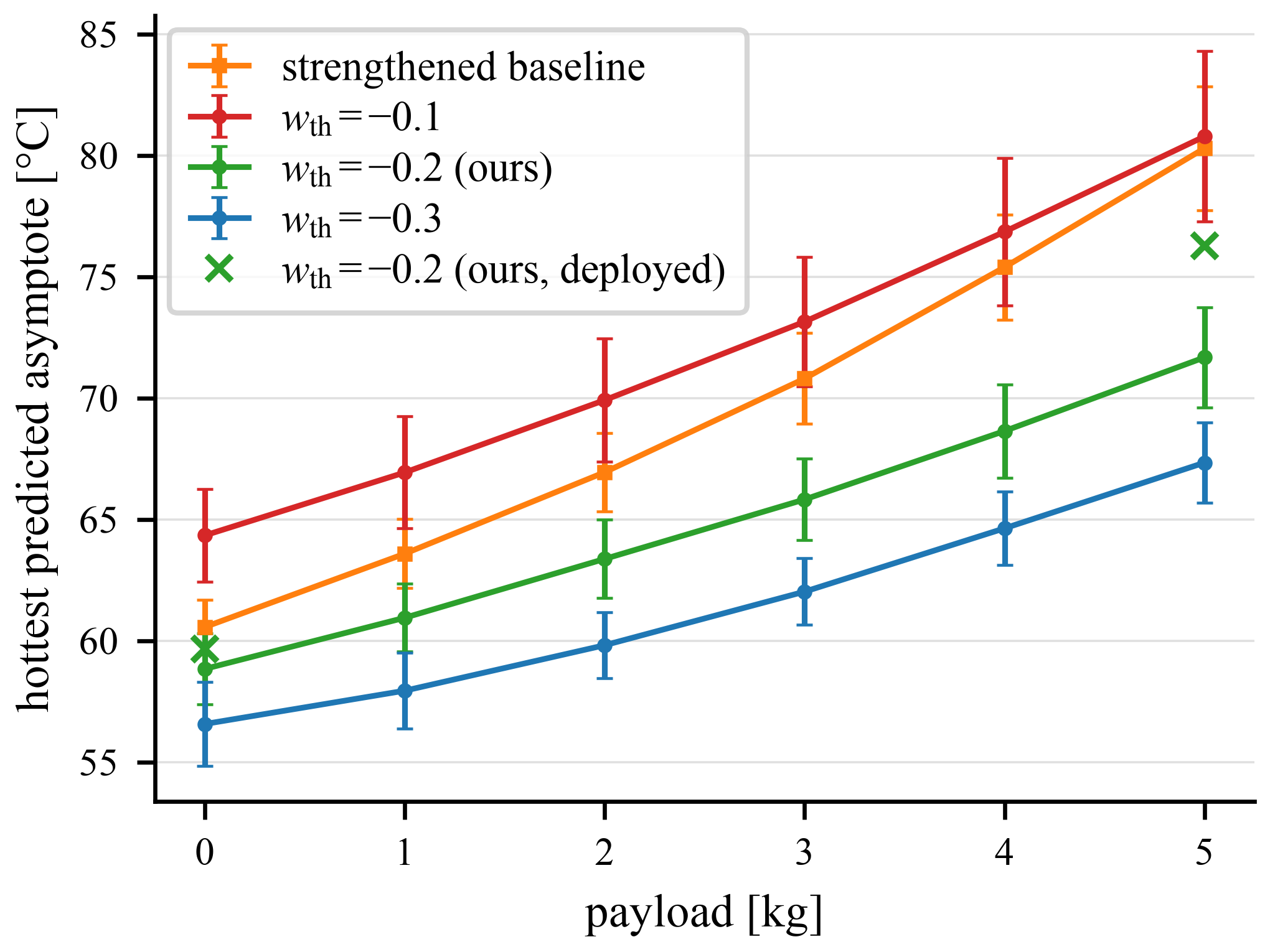}
\par\vspace{-2mm}
\caption{Hottest predicted asymptote versus payload at 1\,m/s. Lines: simulation, means and sample s.d. across five seeds $\times$ five checkpoints. Crosses: the two deployed runs of our policy, $w_{\mathrm{th}}=-0.2$, evaluated at the simulation ambient of 30\,\textdegree C.}
\label{fig:gait-eval}
\vspace{-1mm}
\end{figure}

In Fig.~\ref{fig:gait-eval}, we report the gait evaluation results for $w_{\mathrm{th}}$ of $-0.1$, $-0.2$ and $-0.3$, as well as for the strengthened baseline. The baseline lies off the chart, with hottest predicted asymptotes of $91\pm8$\,\textdegree C unloaded and $131\pm14$\,\textdegree C with a 5\,kg payload, above the threshold even unloaded. The hottest asymptote grows faster than linearly with payload for all policies, but the growth is markedly shallower under the thermal reward. The reward lowers the hottest asymptote at every payload and its weight sets the margin: with 5\,kg, $w_{\mathrm{th}}=-0.2$ and $-0.3$ settle at 72 and 67\,\textdegree C against 80\,\textdegree C for the strengthened baseline, whereas $-0.1$ does not yet beat the squared-torque penalty of the strengthened baseline (claim (iii)).
% \vspace{-1mm}
\subsection{Terrain traversal in simulation} \label{sec:terrain}
% \vspace{-1mm}
% Fig. 5 uses all five checkpoints per run from the full 0.8 m/s campaign, without filtering.
We evaluate the policies of Sec.~\ref{sec:gait-eval} (five seeds $\times$ five checkpoints) on five maximum-difficulty training terrains with payloads of 0, 2.5 and 5\,kg, using 100 rollouts per checkpoint--terrain--payload combination. After a flat approach, each robot receives a body-frame forward/lateral command $c_{\mathrm{lin}}$ for 5\,s: $\|c_{\mathrm{lin}}\|_2=0.8$\,m/s, with a positive forward component and a lateral component evenly spaced over $[-0.2,0.2]$\,m/s. We apply heading control targeting a terrain-orthogonal heading throughout. Survival requires no base-link collision or flipping during approach or scoring.

For survivors, we report two tracking errors:
\begin{itemize}
\item the normalized horizontal endpoint error $E_p=\|p_5-p_{\mathrm{ref}}\|_2/(4\,\mathrm{m})$, where $p_5$ is the horizontal position of the robot after five seconds. The reference endpoint $p_{\mathrm{ref}}$ is computed by moving the actual scoring-start position of the robot for five seconds at the commanded forward and lateral speeds, along fixed horizontal directions aligned with and perpendicular to the target heading, respectively.
\item the mean instantaneous velocity error: the mean Euclidean difference between measured body-frame forward and lateral velocity and the corresponding command $c_{\mathrm{lin}}$ over 250 samples per rollout.
\end{itemize}

\begin{figure}[t]
\vspace{1.5mm}
\centering
\includegraphics[width=0.97\columnwidth]{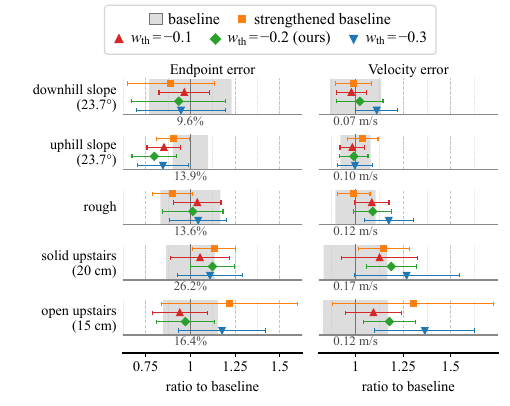}
\par\vspace{-3mm}
\caption{Tracking errors among 5\,s survivors relative to the baseline (1; lower is better). Means and sample s.d. across five seeds $\times$ five checkpoints, after averaging payloads, are divided by the baseline mean. Grey bands show baseline s.d.; numbers below each row give its absolute mean.}
\label{fig:terrain-traversal}
\vspace{-1mm}
\end{figure}

\looseness=-1
Fig.~\ref{fig:terrain-traversal} shows tracking errors relative to the baseline. With $w_{\mathrm{th}}=-0.1$, no mean error increases by more than 0.81 baseline s.d., and the uphill-slope endpoint error drops by 14.8 \%. Our deployed weight, $w_{\mathrm{th}}=-0.2$, cuts that error by 20.2 \% and costs 18--19 \% more velocity error on stairs, about 0.02--0.03\,m/s at a 0.8\,m/s command. The strengthened baseline incurs similar increases in stair velocity error with a far larger spread on open stairs, which suggests the penalty comes with lowering torque rather than being specific to our reward. At $w_{\mathrm{th}}=-0.3$, velocity error increases on stairs, rough terrain, and the downhill slope, for only 2--4\,\textdegree C more margin (Fig.~\ref{fig:gait-eval}), so we deploy $w_{\mathrm{th}}=-0.2$ in Secs.~{\ref{sec:flat-endurance}} and~{\ref{sec:factory-circuit}}. All recipes achieve at least 99.7 \% pooled survival on these five terrains, which supports claim (iv), that our method retains robust terrain traversal ability.

% Keep the circuit collage near its first detailed description.
\begin{figure*}[!t]
\vspace{2mm}
\centering
% Equal panel heights; the fifth panel is an illustrative, non-surveyed model.
\begin{tabular}{@{}c@{\hspace{0.006\textwidth}}c@{\hspace{0.006\textwidth}}c@{\hspace{0.006\textwidth}}c@{\hspace{0.006\textwidth}}c@{}}
\includegraphics[height=0.185\textwidth]{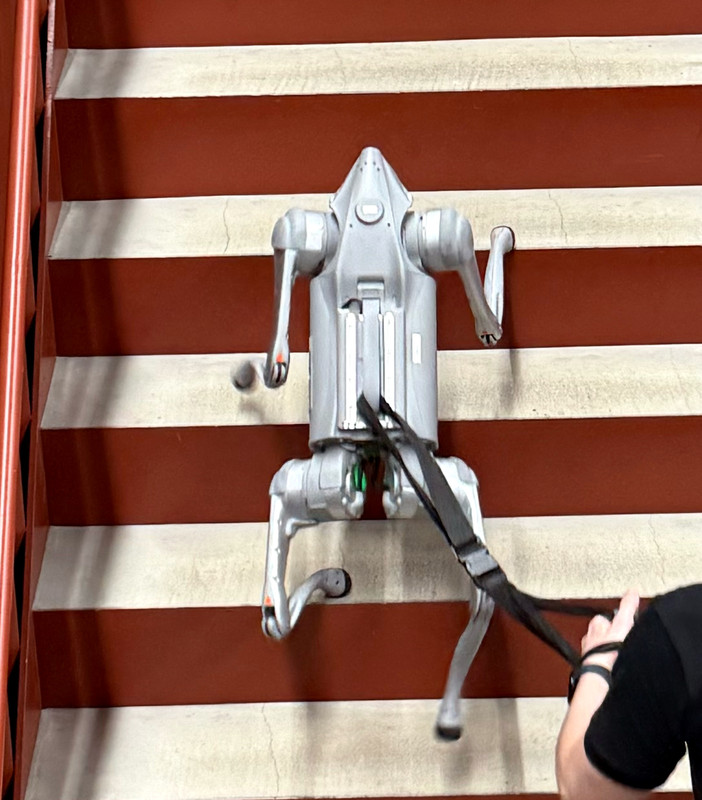} &
\includegraphics[height=0.185\textwidth]{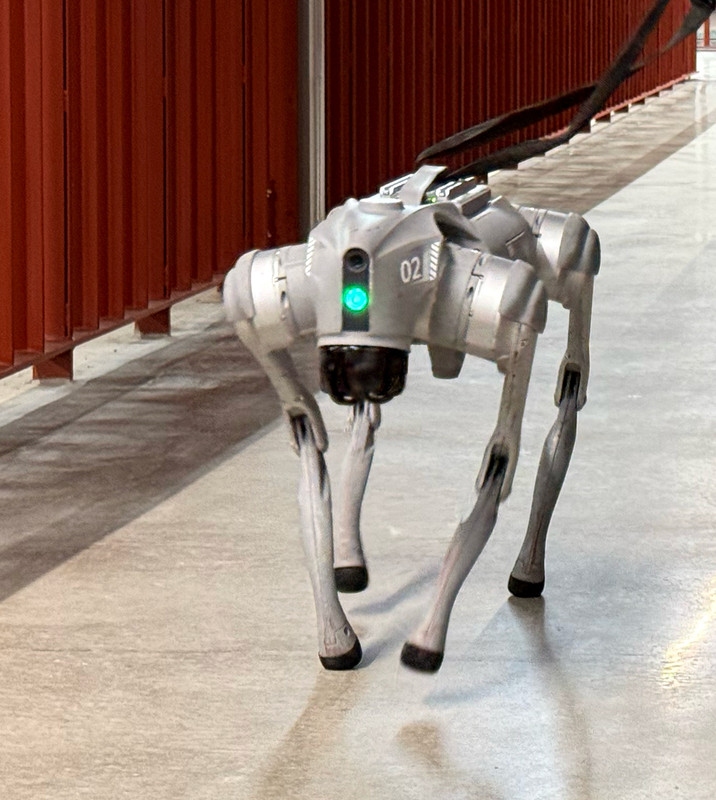} &
\includegraphics[height=0.185\textwidth]{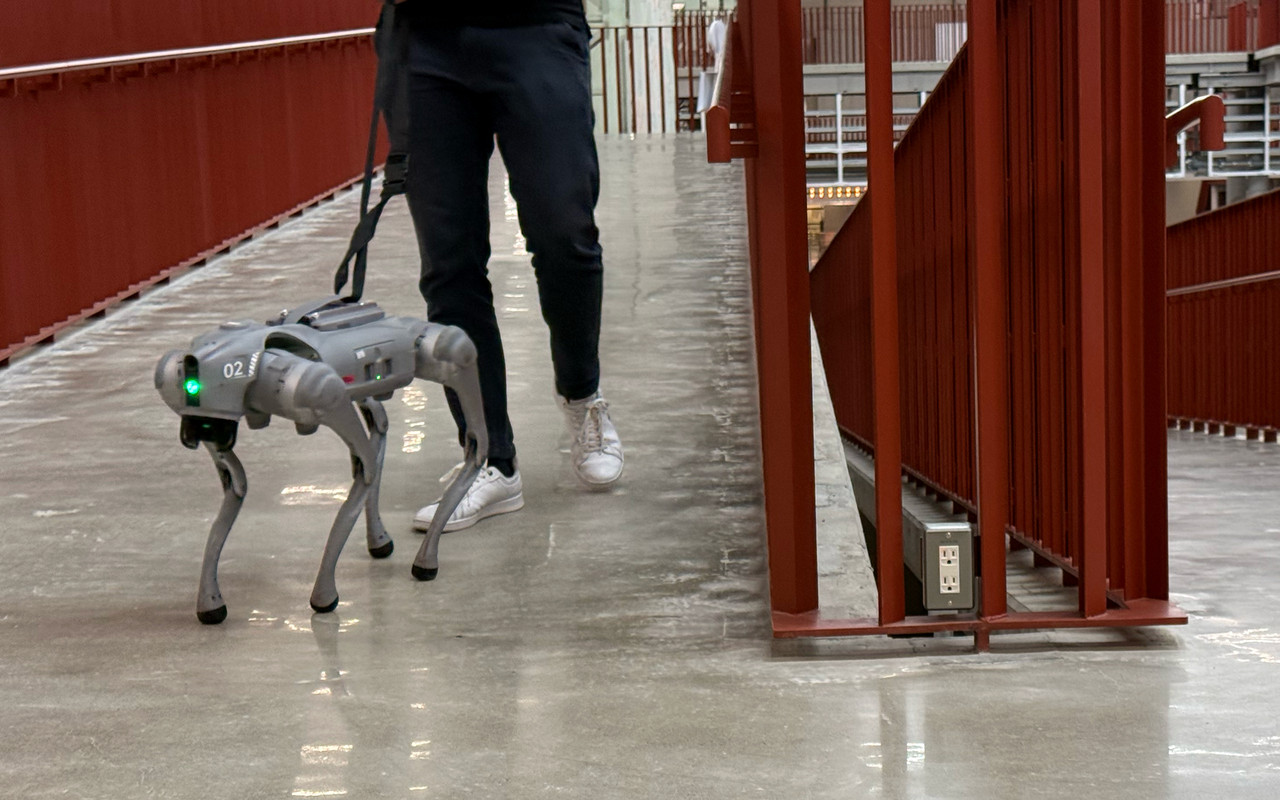} &
\includegraphics[height=0.185\textwidth]{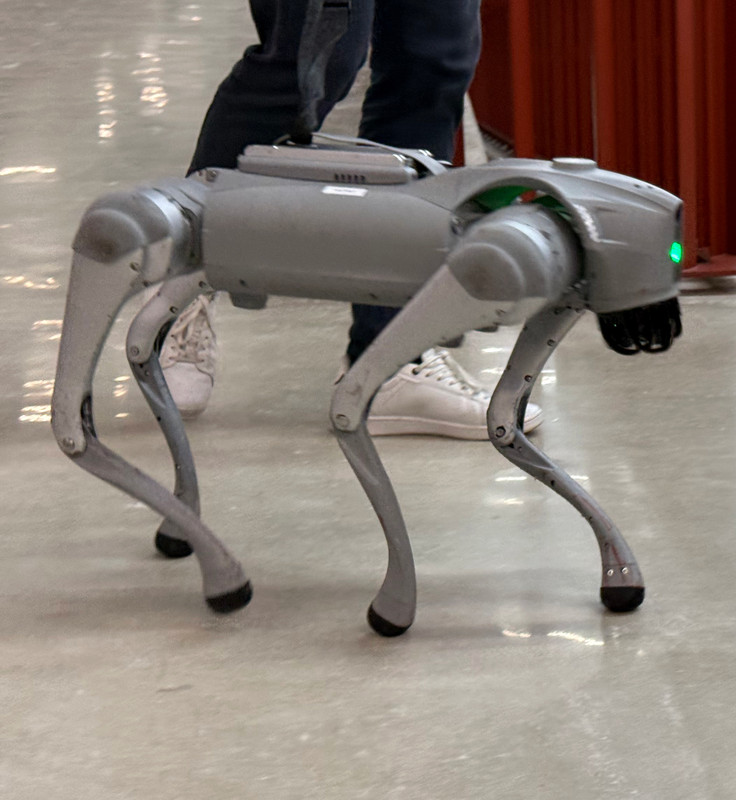} &
\includegraphics[width=0.18\textwidth,height=0.185\textwidth]{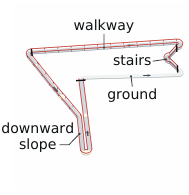} \\[2pt]
\small (a) Stair ascent & \small (b) Walkway & \small (c) Ramp descent & \small (d) Landing & \small (e) Schematic
\end{tabular}
\par\vspace{-2mm}
\caption{Views of the factory circuit, showing the gait of our thermal policy across different terrain types.}
\label{fig:circuit-photos}
\vspace{-1mm}
\end{figure*}

\subsection{Endurance on flat ground}\label{sec:flat-endurance}

We command the baseline and our thermal policy at 1\,m/s with a 5\,kg payload in the same large room as Sec.~\ref{sec:exp-protocol} until thermal shutdown or battery depletion.

While the baseline shuts down in 19 minutes (1.1\,km), our thermal policy lasts 174 minutes (10.0\,km) until battery depletion (Fig.~\ref{fig:baseline-vs-ours}), with 14\,\textdegree C thermal headroom. These runs were at 22 and 26\,\textdegree C room temperature (cooler day for the baseline). Lower Joule losses also make the policy more economical: at the same speed it draws 142\,W against 244\,W for the baseline, an electrical cost of transport of 0.71 against 1.24 (43 \% lower, quiescent power included and Go2 measured mass of 16.2\,kg plus payload). The supplementary video shows the stark contrast in posture: level and upright for our thermal policy, crouched for the baseline.
The 5\,kg run of Fig.~\ref{fig:baseline-vs-ours} and an unloaded two-hour run of the same thermal policy also quantify the sim-to-real gap: the hottest asymptotes predicted from their measured gait statistics (crosses in Fig.~\ref{fig:gait-eval}) exceed the simulated means by 0.8\,\textdegree C unloaded and 4.6\,\textdegree C with the payload.

\vspace{-1mm}
\subsection{Factory circuit}\label{sec:factory-circuit}

\begin{figure}[t]
\centering
\includegraphics[width=0.90\columnwidth]{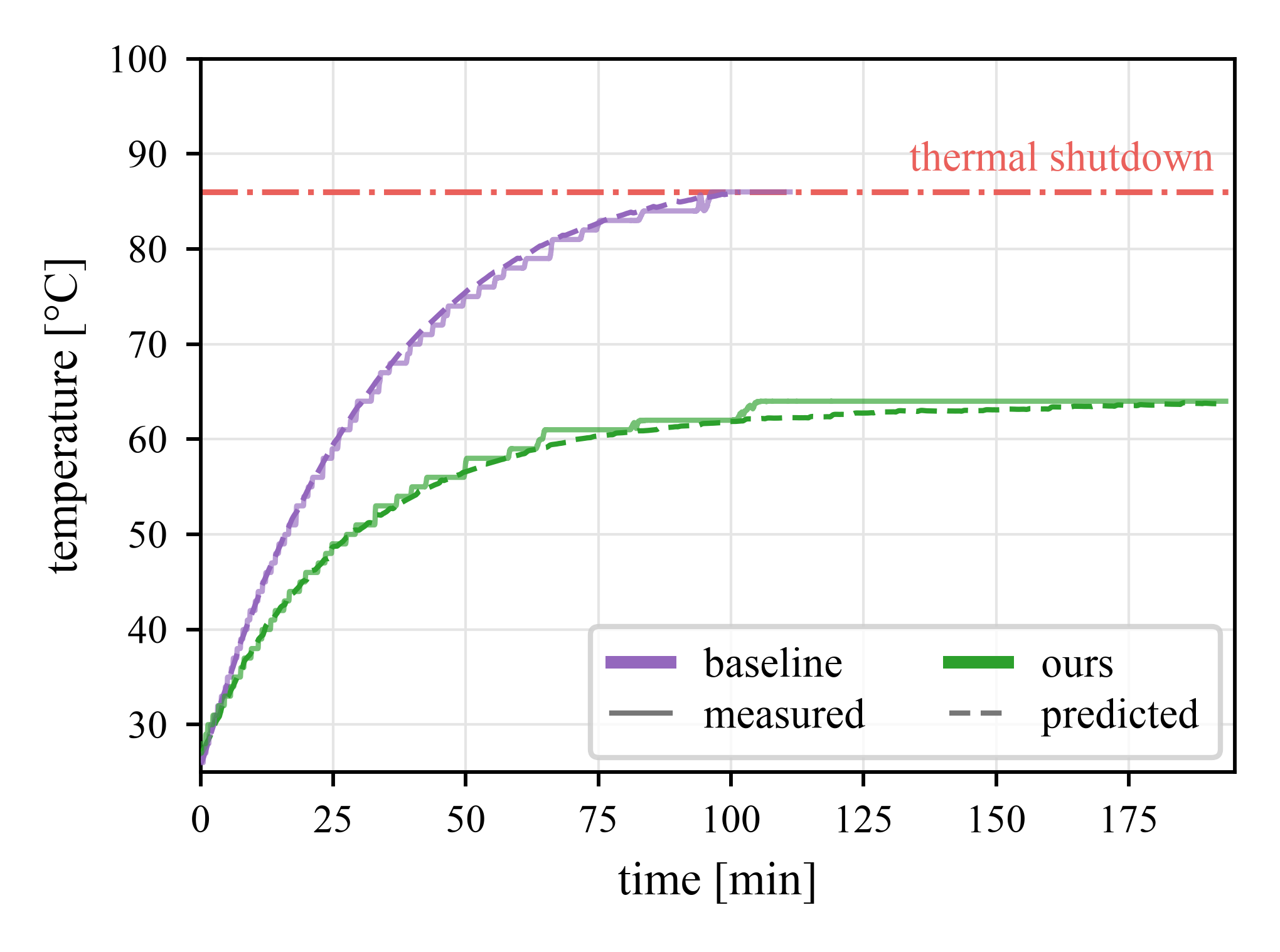}
\par\vspace{-2mm}
\caption{Factory circuit, no payload: hottest joint of the baseline (purple) and our thermal policy (green); measured (solid) and predicted (dashed).}
\label{fig:factory-circuit}
\vspace{-1mm}
\end{figure}

We command our robot along a continuous factory circuit (Fig.~\ref{fig:circuit-photos}) combining a steep staircase climbed upward (19.5\,cm risers with $\pm1$\,cm variability, 25.2\,cm tread depth), a walkway, a gentle downhill ramp (3.2\,\textdegree average slope), flat stretches, and sharp turns (239\,m per lap; 0.7\,m/s velocity command on stairs, 1\,m/s otherwise), until thermal shutdown or battery depletion.

In Fig.~\ref{fig:factory-circuit}, the baseline reaches thermal shutdown after 111 minutes (25 laps, 6.2\,km); our thermal policy lasts 193 minutes (47 laps, 11.2\,km) before its battery is empty, its hottest joint peaking at 64\,\textdegree C (22\,\textdegree C margin). The cost of transport drops from 1.27 to 0.83 (34 \% lower), or 8.8 instead of 13.4\,Wh per lap. The circuit runs were at 23 and 25\,\textdegree C room temperature (cooler for the baseline).

On hardware, our policy thus traverses the same stairs and ramps as the baseline but lasts a full battery with 22\,\textdegree C of margin (claims (iv) and (v)).

% \vspace{-1mm}
\section{Conclusion}
% \vspace{-0.5mm}

We have shown that a learned locomotion policy can be made thermally aware without observing temperature. Instead of penalizing temperatures, which barely move within an episode, our reward penalizes the thermal steady states the current gait is heading toward: with an identified thermal network of the robot, they follow from gait-averaged squared torques and joint speeds through one small linear solve and respond to the policy within a gait cycle. The 22-parameter network of the Unitree Go2 behind it, calibrated against the measured electrical power, predicts all twelve motor temperatures with a mean error below 1\,\textdegree C on held-out runs across robots, gaits, payloads, and speeds. On hardware, the Go2 went from temperature-limited to battery-limited: 174 instead of 19 minutes at 1\,m/s with 5\,kg of payload, with 14\,\textdegree C of thermal headroom, and 47 laps of a factory circuit with steep stairs and ramps on one battery, where the baseline overheated after 25, and with no loss of terrain traversal capability. In future work, we plan to (1) close the sim-to-real gap we observed at large payloads, (2) build an electrical power reward on the identified power balance to turn thermal headroom into range, and (3) add a governor that adjusts the commanded velocity from the model's real-time predictions.

\section*{Data Availability}

All real-robot recordings used in this paper are openly available~\cite{segas2026zenodo}: the two training runs, the nine validation runs, and the twelve test runs (totaling 48 hours on two robots). Each recording provides the twelve motor temperatures, joint torque and velocity statistics, and bus voltage and current every 2\,s, together with the ambient temperature and metadata (e.g., payload, measured speeds). These data suffice to re-identify the thermal network of Tab.~\ref{tab:parameters} and to reproduce Tab.~\ref{tab:thermal-performance}.

\section*{Acknowledgment}

We are very grateful to Mutsumi Iizuka for designing and procuring the payload and its mounting. We thank Ikuo Tachi and Kenta Takeshima for making our experiments at Woven City's Inventor Garage possible and for accommodating our tight schedule, and Takato Seki and Yutaka Takaoka for their support of this project.

% References are maintained in the project-local bibliography.
\bibliographystyle{plain}
\bibliography{new}

\useRomanappendicesfalse
\appendices

\begin{table}[!b]
\caption{All 22 identified parameters.}
\label{tab:parameters}
\vspace{-2mm}
\centering
\footnotesize
\renewcommand{\arraystretch}{0.94}
\setlength{\tabcolsep}{4pt}
\scalebox{0.94}{%
\begin{tabular}{@{}l l r l@{}}
\toprule
Parameter & Symbol & Value & Unit \\
\midrule
\multicolumn{4}{@{}l}{\textbf{Electrically identified}} \\
Total quiescent power & $p^{\mathrm{tot}}_{\idle}$ & 43.9 & W \\
Joule heating coefficient & $k_m$ & 0.494 & W/(N\,m)$^2$ \\
Resistance temp.\ coefficient & $\alpha$ & $3.93\cdot10^{-3}$ & K$^{-1}$ \\
\midrule
\multicolumn{4}{@{}l}{\textbf{Thermally identified: heat sources}} \\
Calf bypass share to body & $\varphi_{\mathrm{calf}}$ & 0.29 & -- \\
Friction coefficient & $f$ & 0.244 & W/(rad/s) \\
Actuator quiescent power & $p_{\idle}$ & 1.90 & W \\
\addlinespace[3pt]
\multicolumn{4}{@{}l}{\textbf{Thermally identified: cooling to ambient}} \\
Hip to ambient & $g^{a}_{\mathrm{hip}}$ & 0.079 & W/K \\
Thigh and calf to ambient & $g^{a}_{\mathrm{tc}}$ & 0.180 & W/K \\
Body to ambient & $g_b$ & 6.3 & W/K \\
Forced-convection gain & $c_v$ & 0.50 & (s/m)$^{n}$ \\
Forced-convection exponent & $n$ & 0.68 & -- \\
Rear-wake factor & $s_{\mathrm{rear}}$ & 0.69 & -- \\
\addlinespace[3pt]
\multicolumn{4}{@{}l}{\textbf{Thermally identified: conduction edges}} \\
\textbf{Calf--thigh} & $g_{\mathrm{thigh,calf}}$ & \textbf{0.335} & W/K \\
Hip--thigh & $g_{\mathrm{hip,thigh}}$ & 0.040 & W/K \\
Hip--body, front legs & $g^{B}_{\mathrm{hip,F}}$ & 0.114 & W/K \\
Hip--body, rear legs & $g^{B}_{\mathrm{hip,R}}$ & 0.080 & W/K \\
Thigh--body, front legs & $g^{B}_{\mathrm{thigh,F}}$ & 0.067 & W/K \\
Thigh--body, RR leg & $g^{B}_{\mathrm{thigh,RR}}$ & 0.127 & W/K \\
Thigh--body, RL leg & $g^{B}_{\mathrm{thigh,RL}}$ & 0.217 & W/K \\
\addlinespace[3pt]
\multicolumn{4}{@{}l}{\textbf{Thermally identified: capacities and lag}} \\
Actuator heat capacity & $C_m$ & 421 & J/K \\
Body heat capacity & $C_B$ & 18.1 & kJ/K \\
Observation lag & $\tau_{\obs}$ & 12.2 & s \\
\bottomrule
\end{tabular}%
}
\par\vspace{-2mm}
\end{table}

\section{Full node-level equations} \label{app:node-level}

With the heat sources \eqref{eq:copper}--\eqref{eq:sources}, each actuator node obeys
\begin{equation}
\begin{aligned}
C_m\,\dot T_j &\;=\; Q_j
\;-\; g^{a}_{y(j)}\bigl(1+c_v\,s_{\ell(j)}\,v^{\,n}\bigr)\,(T_j-T_a)
\\
&\;-\!\!\sum_{k\in\mathcal{N}(j)}\!\! g_{jk}\,(T_j-T_k)
\;-\; g^{B}_{j}\,(T_j-T_B).
\end{aligned}
\label{eq:node}
\end{equation}
Each node shares heat with its neighboring nodes $\mathcal{N}(j)$ (identified conduction topology, Tab.~\ref{tab:parameters}). Cooling to ambient uses a natural-convection conductance per actuator type---$g^{a}_{\mathrm{hip}}$ for the hips located deep within the torso, and one shared $g^{a}_{\mathrm{tc}}$ for the closely located thigh and calf units---enhanced by power-law forced convection $c_v v^{n}$ and attenuated for rear legs by the rear-wake (shielding) factor $s_\ell$ ($s_\ell=s_{\mathrm{rear}}$ for rear legs, $1$ for front).

The hidden body node has the same structure:
\begin{equation}
\begin{aligned}
C_B\,\dot T_B &\;=\; Q^{B}
\;-\; g_b\,(T_B-T_a)
\\
&\;-\; \sum_{j=1}^{12} g^{B}_{j}\,(T_B-T_j),
\end{aligned}
\label{eq:body}
\end{equation}
and the displayed temperature $\hat T_j$ lags the node state:
\begin{equation}
\tau_{\obs}\,\dot{\hat T}_j \;=\; T_j-\hat T_j.
\label{eq:lag}
\end{equation}

Stacking \eqref{eq:node} and \eqref{eq:body} yields the matrix-vector form \eqref{eq:network}, with initial conditions $T_B(0)=T_a(0)$ and $T_j(0)$ from the first logged samples.

\section{Identified thermal and electrical parameters} \label{app:sys-id}

\emph{Identification procedure.} Averaged over the interior of a segment (first and last 20\,s dropped), \eqref{eq:balance} is linear in the unknowns (segment averaging is omitted in the following for readability). $p^{\mathrm{tot}}_{\idle}$ is directly the mean bus power of the damp segments of both training runs. With $p^{\mathrm{tot}}_{\idle}$ fixed, $k_m$ is the slope of a one-parameter weighted linear least-squares fit, with $\alpha$ held at the copper value, of the Joule power $P_{\mathrm{bus}}-p^{\mathrm{tot}}_{\idle}-\sum_j{\tau_j\dot q_j}$ against $\sum_j r_j {u_j (1+\alpha(\hat T_j-T_{\rmref}))}$ over the 15 static and walking segments of both training runs (weights $1/\sigma$, $\sigma=1$\,W static, 3\,W walking), giving $k_m=0.494\pm0.003$ W/(N\,m)$^2$ with 1.4\,W (static) and 3.4\,W (walking) residuals. The temperature coefficient $\alpha$ follows from the two-regressor extension of the same fit, in which the terms $\sum_j r_j u_j$ and $\sum_j r_j {u_j(\hat T_j-T_{\rmref})}$ are separated; the cold and hot back-splayed stances of protocol A provide its largest leverage. It yields $\alpha=(3.5\pm1.1)\cdot10^{-3}$ K$^{-1}$, and we fix $\alpha$ at the copper value $3.93\cdot10^{-3}$ K$^{-1}$, which lies inside that interval.
We obtain the remaining 19 parameters by nonlinear least squares. For a candidate parameter vector we integrate \eqref{eq:network} with an implicit Euler scheme at the 2 s logging period over both training runs, from the initial conditions of Appendix~\ref{app:node-level}, and minimize the unweighted sum of squared differences between the lagged model temperatures $\hat T_j$ and the displayed temperatures over all samples and all twelve joints (about $1.9\cdot10^{5}$ residuals) with a trust-region reflective solver. We optimize positive parameters in log space, and one fit takes a few seconds; four restarts perturbed by $\pm30$ \% return the same optimum.

\emph{Identified parameters.} Tab.~\ref{tab:parameters} lists all 22 identified parameters (3 electrical + 19 thermal). The board quiescent power follows as $p^{B}_{\idle}=p^{\mathrm{tot}}_{\idle}-12\,p_{\idle}=21.1$\,W. We free the per-type bypass shares $\varphi_{\mathrm{hip}}$ and $\varphi_{\mathrm{thigh}}$ during structure selection; they converge to zero, so we do not count them. The identified topology has no conduction edge between legs and no calf--body edge; the calf--thigh edge (bold) dominates, and body edges differ front/rear and, for the rear thighs, left/right, since the battery pack (the largest lumped mass in the torso) sits closest to the rear-left leg.

\end{document}